%% file: mdpp.tex
\documentclass[twoside]{article}
\usepackage[accepted]{aistats2016}
\usepackage{natbib}
\usepackage{amsmath}
\usepackage{amssymb}
\usepackage{hyperref}
\usepackage{url}
\usepackage[]{graphicx}
\usepackage{verbatim}
\usepackage{listings}
\usepackage{rotating}
\usepackage[scaled=0.95]{inconsolata}

\usepackage{microtype}

\usepackage{caption}
\usepackage{algorithm}
\usepackage{algpseudocode}

\input{macros.tex}
\input{anglican.tex}

\begin{document}

\twocolumn[

\aistatstitle{Black-Box Policy Search with Probabilistic Programs}

\aistatsauthor{Jan-Willem van de Meent \And Brooks Paige \And David Tolpin \And Frank Wood }

\aistatsaddress{ Department of Engineering Science, University of Oxford} ]

\begin{abstract}
\input{abstract}

\end{abstract}

\section{Introduction}

\input{introduction}
\begin{figure*}[!t]
\begin{minipage}[t]{0.5\textwidth}

\begin{lstlisting}[frame=bottomline, firstline=47, lastline=64]
(defquery ctp
 "Probabilistic program representing an agent 
  solving the Canadian Traveler Problem"
 [graph src tgt base-prob make-policy]
 (let [sub-graph 
        (sample-weather graph base-prob src tgt)
       [path dist counts] 
        (dfs-agent sub-graph src tgt (make-policy))]
  (factor (- dist))
  (predict :path path)
  (predict :distance dist)
  (predict :counts counts)))

(defm dfs-agent
 "Run depth-first-search from start to target,
  prioritizing edges according to policy"
 [graph start target policy]
   ... ) 
\end{lstlisting}
\end{minipage}
~~~
\begin{minipage}[t]{0.45\textwidth}
\begin{lstlisting}[frame=bottomline, firstline=1, lastline=18]
(defm make-random-policy 
 "Policy: Select edge at random"
 []
 (fn policy [u vs]
  (sample 
   (categorical 
    (zipmap vs (repeat (count vs) 1.))))))

(defm make-edge-policy
 "Policy: learn priorities for each edge"
 []
 (let [Q (mem (fn [u v]
               (sample [u v] 
                (tag :policy
                 (gamma 1. 1.)))))]
  (fn policy [u vs]
   (argmax 
    (zipmap vs (map (fn [v] (Q u v)) vs))))))
\end{lstlisting}
\end{minipage}
~~~
\begin{minipage}[t]{\textwidth}
\includegraphics[width=1.6in,trim={0.2in 0.3in 0 0.1in},clip]{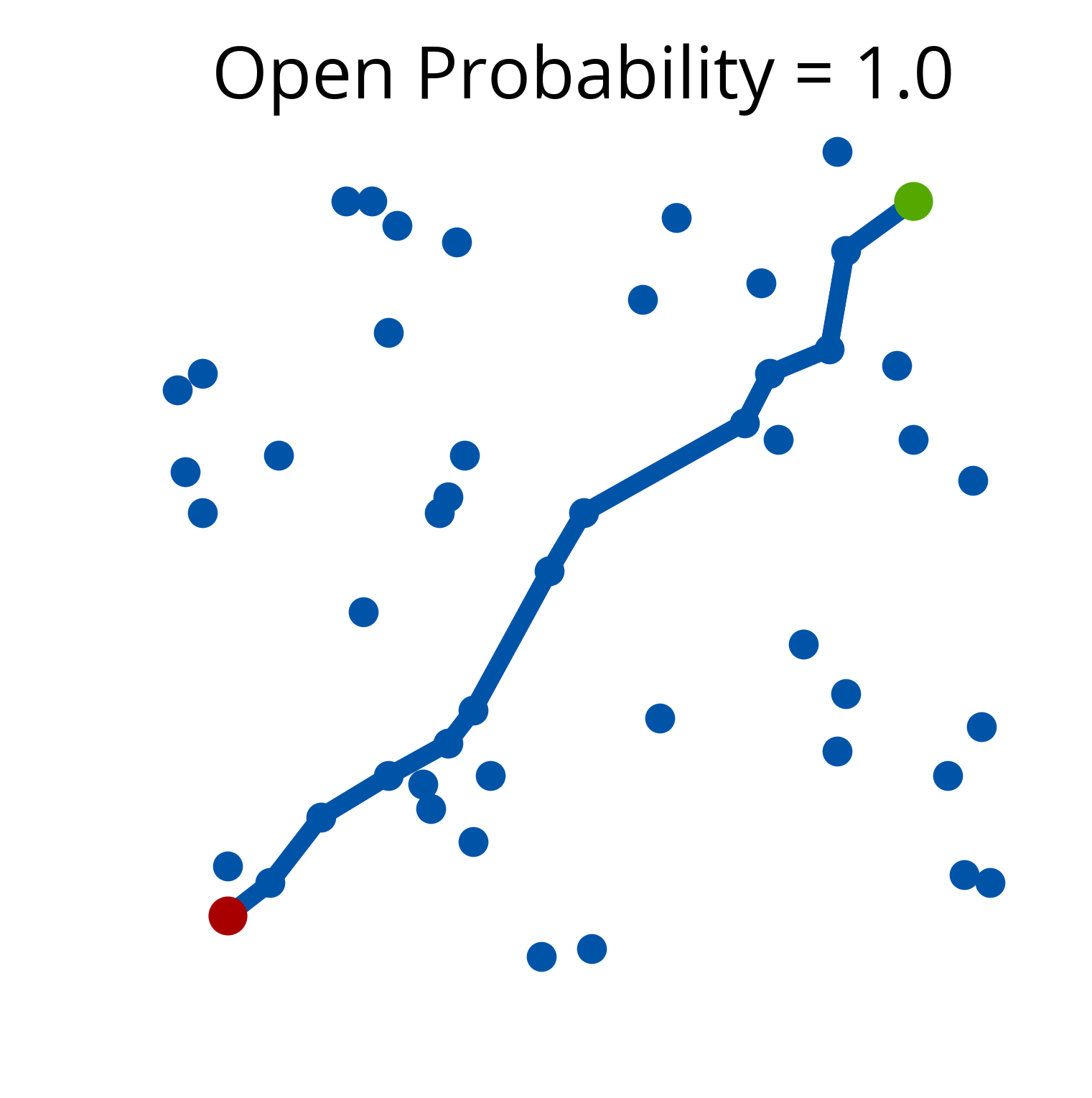}
\hspace{0.055in}
\includegraphics[width=1.6in,trim={0.2in 0.3in 0 0.1in},clip]{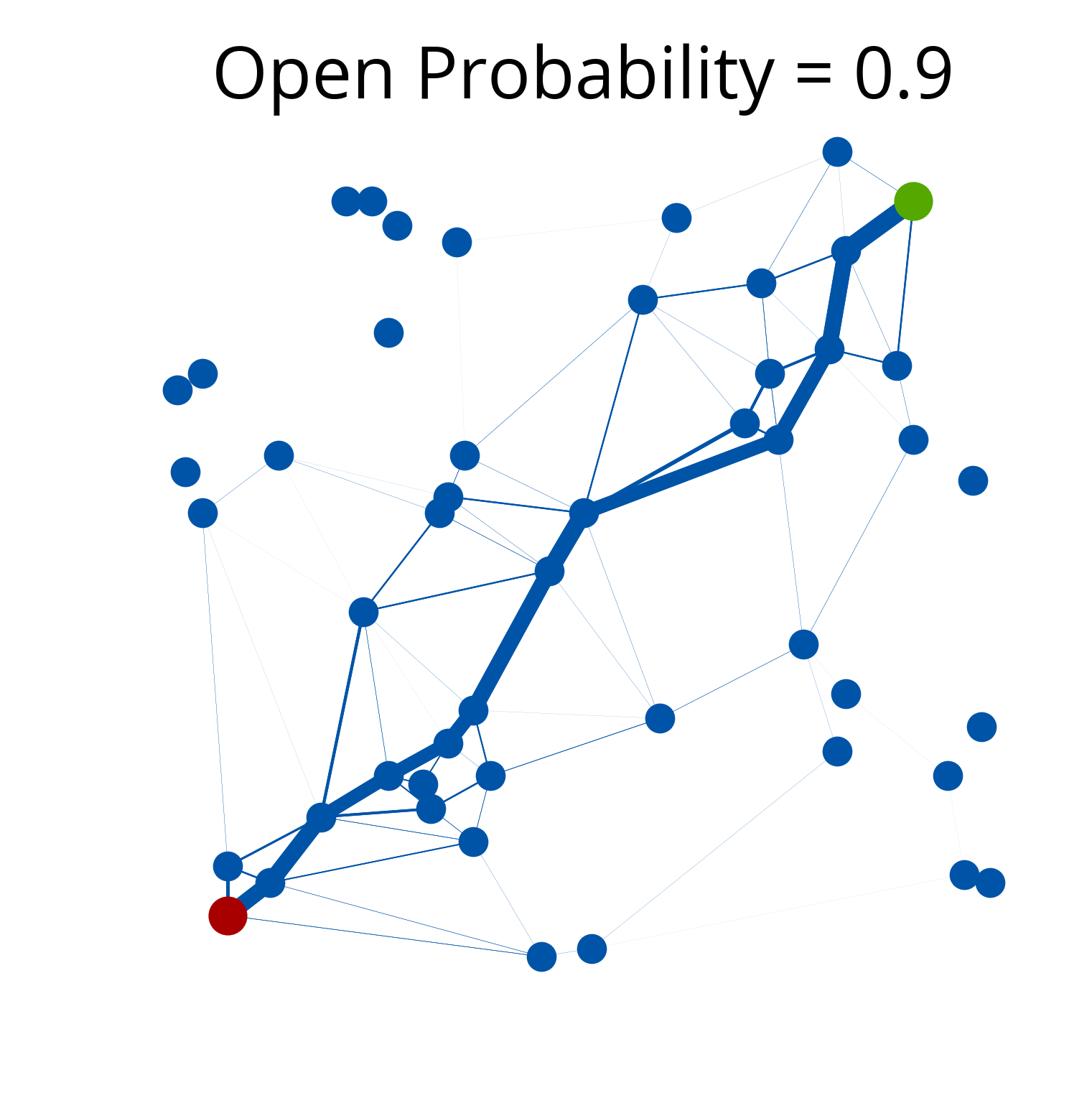}
\hspace{0.055in}
\includegraphics[width=1.6in,trim={0.2in 0.3in 0 0.1in},clip]{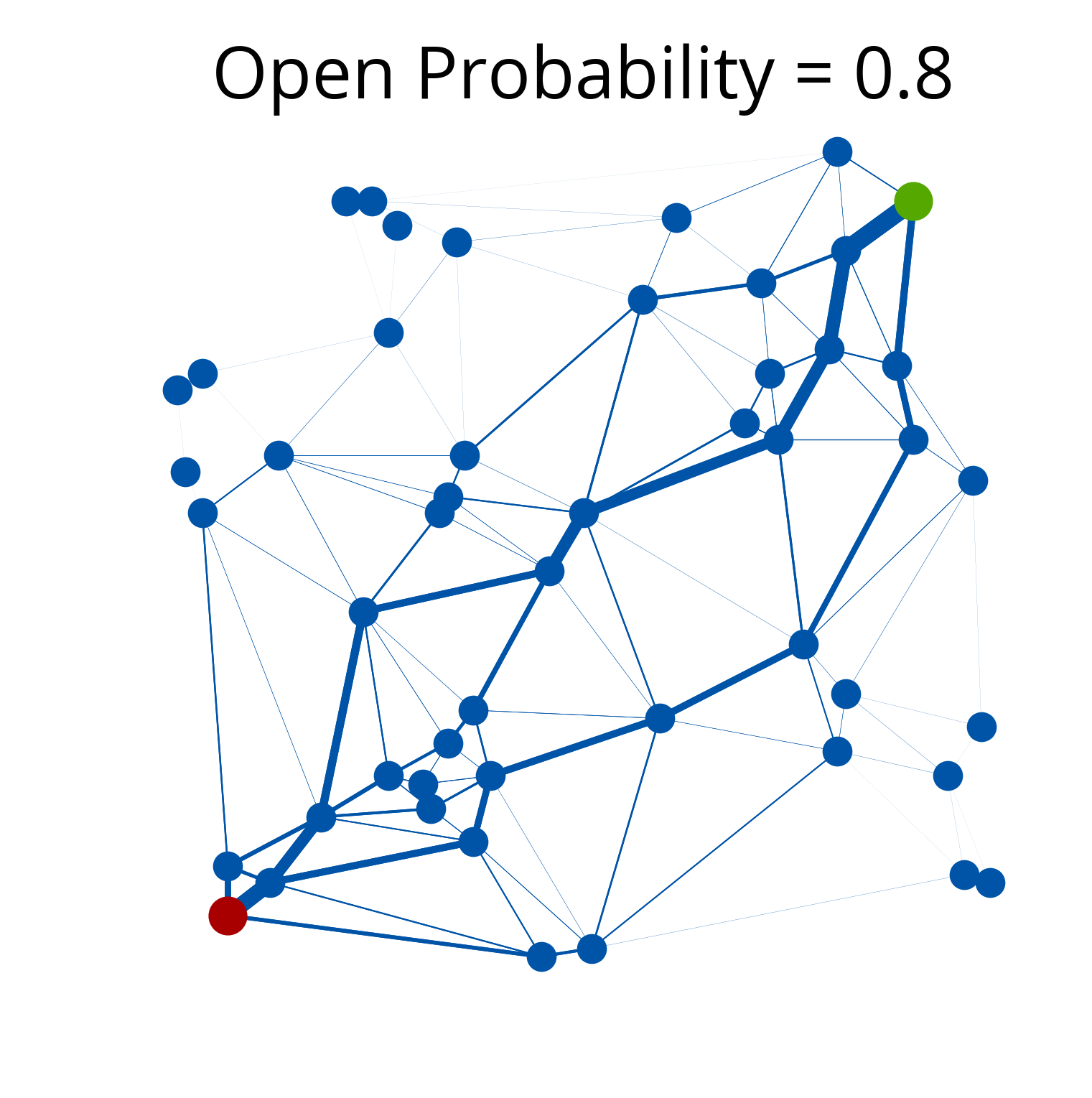}
\hspace{0.055in}
\includegraphics[width=1.6in,trim={0.2in 0.3in 0 0.1in},clip]{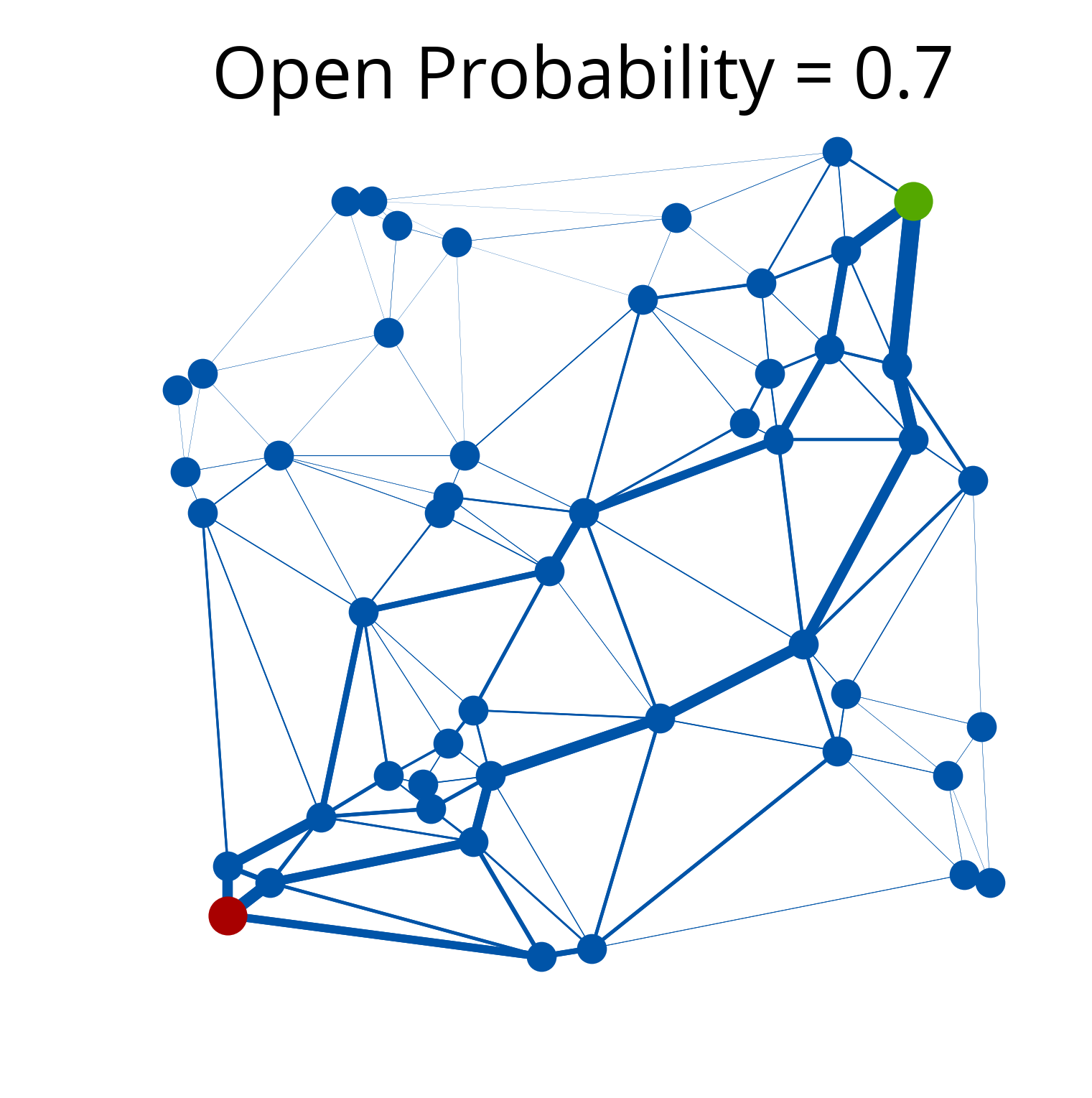}
\end{minipage}
\caption{\label{fig:ctp-overview} A Canadian traveler problem (CTP) implementation in Anglican. In the CTP, an  agent must travel along a graph, which represents a network of roads, to get from the start node (green) to the target node (red). Due to bad weather some roads are blocked, but the agent does not know which in advance.
Upon arrival at each node the agent observes the set of open edges. The function \lsi{dfs-agent} walks the graph by performing depth-first search, calling a function \lsi{policy} to choose the next destination based on the current and unvisited locations. The function \lsi{make-random-policy} returns a \lsi{policy} function that selects destinations uniformly at random, whereas \lsi{make-edge-policy} constructs a \lsi{policy} that selects according to sampled edge preferences \lsi{(Q u v)}. By learning a distribution on each value \lsi{(Q u v)} through gradient ascent on the marginal likelihood, we obtain a heuristic offline policy that follows the shortest path when all edges are open, and explores more alternate routes as more edges are closed.}

\end{figure*}


\input{background}

\section{Black-box Policy Search}

\input{methodology}

\section{Learning Probabilistic Programs}

\input{programs}



\section{Case Studies}

\input{studies}

\section{Discussion}

\input{discussion}

\input{acknowledgements}

\small
\bibliographystyle{abbrvnat}
\bibliography{mdpp}

\newpage
\onecolumn
\appendix
\input{appendix.tex}

\end{document}

%% file: macros.tex
\newcommand{\p}[2]{\ensuremath{p(#1 \,|\, #2)}}
\newcommand{\q}[2]{\ensuremath{q(#1 \,|\, #2)}}

\newcommand{\x}{\ensuremath{x}}
\newcommand{\y}{\ensuremath{y}}
\newcommand{\z}{\ensuremath{z}}
\renewcommand{\u}{\ensuremath{u}}
\renewcommand{\q}{\ensuremath{\theta}}
\renewcommand{\l}{\ensuremath{\lambda}}
\renewcommand{\t}{\ensuremath{\tau}}

\newcommand{\Dkl}[2]{\ensuremath{D_{\rm KL}(#1 \,||\, #2)}}
\newcommand{\E}{\ensuremath{\mathbb{E}}}
\renewcommand{\L}{\ensuremath{\mathcal{L}}}
\newcommand{\Q}{\ensuremath{\mathcal{Q}}}
\renewcommand{\P}{\ensuremath{{\mathcal P}}}
\newcommand{\B}{\ensuremath{{\mathcal B}}}
\newcommand{\R}{\ensuremath{\mathbb{R}}}

\usepackage{xcolor}

%% file: anglican.tex
\usepackage{listings}
\usepackage{color}
\usepackage{xcolor}
\definecolor{blue}{rgb}{0,0.33,0.66}
\definecolor{red}{rgb}{0.66,0.0,0.0}
\definecolor{purple}{rgb}{0.33,0,0.66}
\definecolor{cyan}{rgb}{0.0,0.5,0.5}
\definecolor{orange}{rgb}{0.7,0.5,0.0}
\definecolor{gray}{rgb}{0.4,0.4,0.4}

\newcommand{\lsi}[1]{\lstinline$#1$}

\lstdefinelanguage{anglican}%
{
 morekeywords=[1]{->, ->>, .., amap, and, areduce, as->, assert, binding, %
   bound-fn, case, comment, cond, cond->, cond->>, condp, declare, definline, %
   definterface, defmacro, defmethod, defmulti, defn, defn-, defonce, %
   defprotocol, defrecord, defstruct, deftype, delay, doseq, dosync, dotimes, %
   doto, extend-protocol, extend-type, fn, for, future, gen-class, %
   gen-interface, if, if-let, if-not, if-some, import, io!, lazy-cat, lazy-seq, let, %
   letfn, locking, loop, memfn, ns, or, proxy, proxy-super, pvalues, %
   recur, refer-clojure, reify, some->, some->>, sync, time, when, when-first, %
   when-let, when-not, when-some, while, with-bindings, with-in-str, %
   with-loading-context, with-local-vars, with-open, with-out-str, %
   with-precision, with-redefs}, %
  morekeywords=[2]{*, *', +, +', -, -', ->ArrayChunk, ->Vec, ->VecNode, %
  ->VecSeq, -cache-protocol-fn, -reset-methods, /, <, <=, =, ==, >, >=, %
  accessor, aclone, add-classpath, add-watch, agent, agent-error, %
  agent-errors, aget, alength, alias, all-ns, alter, alter-meta!, %
  alter-var-root, ancestors, apply, array-map, aset, aset-boolean, aset-byte, %
  aset-char, aset-double, aset-float, aset-int, aset-long, aset-short, assoc, %
  assoc!, assoc-in, associative?, atom, await, await-for, await1, bases, bean, %
  bigdec, bigint, biginteger, bit-and, bit-and-not, bit-clear, bit-flip, %
  bit-not, bit-or, bit-set, bit-shift-left, bit-shift-right, bit-test, %
  bit-xor, boolean, boolean-array, booleans, bound-fn*, bound?, butlast, byte, %
  byte-array, bytes, cast, char, char-array, char?, chars, chunk, %
  chunk-append, chunk-buffer, chunk-cons, chunk-first, chunk-next, chunk-rest, %
  chunked-seq?, class, class?, clear-agent-errors, clojure-version, coll?, %
  commute, comp, comparator, compare, compare-and-set!, compile, complement, %
  concat, conj, conj!, cons, constantly, construct-proxy, contains?, count, %
  counted?, create-ns, create-struct, cycle, dec, dec', decimal?, delay?, %
  deliver, denominator, deref, derive, descendants, destructure, disj, disj!, %
  dissoc, dissoc!, distinct, distinct?, doall, dorun, double, double-array, %
  doubles, drop, drop-last, drop-while, empty, empty?, ensure, %
  enumeration-seq, error-handler, error-mode, eval, even?, every-pred, every?, %
  ex-data, ex-info, extend, extenders, extends?, false?, ffirst, file-seq, %
  filter, filter-ns-publics, filterv, find, find-keyword, find-ns, %
  find-protocol-impl, find-protocol-method, find-var, first, flatten, float, %
  float-array, float?, floats, flush, fn?, fnext, fnil, force, format, %
  frequencies, future-call, future-cancel, future-cancelled?, future-done?, %
  future?, gensym, get, get-in, get-method, get-proxy-class, %
  get-thread-bindings, get-validator, group-by, hash, hash-combine, hash-map, %
  hash-ordered-coll, hash-set, hash-unordered-coll, identical?, identity, %
  ifn?, in-ns, inc, inc', init-proxy, instance?, int, int-array, integer?, %
  interleave, intern, interpose, into, into-array, ints, isa?, iterate, %
  iterator-seq, juxt, keep, keep-indexed, key, keys, keyword, keyword?, last, %
  line-seq, list, list*, list?, load, load-file, load-reader, load-string, %
  loaded-libs, long, long-array, longs, macroexpand, macroexpand-1, %
  make-array, make-hierarchy, map, map-indexed, map?, mapcat, mapv, max, %
  max-key, memoize, merge, merge-with, meta, method-sig, methods, min, %
  min-key, mix-collection-hash, mod, munge, name, namespace, namespace-munge, %
  neg?, newline, next, nfirst, nil?, nnext, not, not-any?, not-empty, %
  not-every?, not=, ns-aliases, ns-functions, ns-imports, ns-interns, %
  ns-macros, ns-map, ns-name, ns-publics, ns-refers, ns-resolve, ns-unalias, %
  ns-unmap, nth, nthnext, nthrest, num, number?, numerator, object-array, %
  odd?, parents, partial, partition, partition-all, partition-by, pcalls, %
  peek, persistent!, pmap, pop, pop!, pop-thread-bindings, pos?, pr, pr-str, %
  prefer-method, prefers, print, print-ctor, print-simple, print-str, printf, %
  println, println-str, prn, prn-str, promise, proxy-call-with-super, %
  proxy-mappings, proxy-name, push-thread-bindings, quot, rand, rand-int, %
  rand-nth, range, ratio?, rational?, rationalize, re-find, re-groups, %
  re-matcher, re-matches, re-pattern, re-seq, read, read-line, read-string, %
  realized?, record?, reduce, reduce-kv, reduced, reduced?, reductions, ref, %
  ref-history-count, ref-max-history, ref-min-history, ref-set, refer, %
  release-pending-sends, rem, remove, remove-all-methods, remove-method, %
  remove-ns, remove-watch, repeat, repeatedly, replace, replicate, require, %
  reset!, reset-meta!, resolve, rest, restart-agent, resultset-seq, reverse, %
  reversible?, rseq, rsubseq, satisfies?, second, select-keys, send, send-off, %
  send-via, seq, seq?, seque, sequence, sequential?, set, %
  set-agent-send-executor!, set-agent-send-off-executor!, set-error-handler!, %
  set-error-mode!, set-validator!, set?, short, short-array, shorts, shuffle, %
  shutdown-agents, slurp, some, some-fn, some?, sort, sort-by, sorted-map, %
  sorted-map-by, sorted-set, sorted-set-by, sorted?, special-symbol?, spit, %
  split-at, split-with, str, string?, struct, struct-map, subs, subseq, %
  subvec, supers, swap!, symbol, symbol?, take, take-last, take-nth, %
  take-while, test, the-ns, thread-bound?, to-array, to-array-2d, trampoline, %
  transient, tree-seq, true?, type, unchecked-add, unchecked-add-int, %
  unchecked-byte, unchecked-char, unchecked-dec, unchecked-dec-int, %
  unchecked-divide-int, unchecked-double, unchecked-float, unchecked-inc, %
  unchecked-inc-int, unchecked-int, unchecked-long, unchecked-multiply, %
  unchecked-multiply-int, unchecked-negate, unchecked-negate-int, %
  unchecked-remainder-int, unchecked-short, unchecked-subtract, %
  unchecked-subtract-int, underive, unsigned-bit-shift-right, update-in, %
  update-proxy, use, val, vals, var-get, var-set, var?, vary-meta, vec, %
  vector, vector-of, vector?, with-bindings*, with-meta, with-redefs-fn, %
  xml-seq, zero?, zipmap}, %
  morekeywords=[3]{factor,observe,predict,sample}, %
  morekeywords=[4]{anglican, cps-fn, def-cps-fn, defanglican, defdist, defm, defquery, %
    doquery, defun, fm, lambda, mem, query, with-primitive-procedures}, %
  morekeywords=[5]{->CRP-process, ->DP-process, ->GP-process, %
  ->bernoulli-distribution, ->beta-distribution, ->binomial-distribution, %
  ->categorical-crp-distribution, ->categorical-distribution, %
  ->categorical-dp-distribution, ->chi-squared-distribution, %
  ->dirichlet-distribution, ->discrete-distribution, %
  ->exponential-distribution, ->flip-distribution, ->gamma-distribution, %
  ->mvn-distribution, ->normal-distribution, ->poisson-distribution, %
  ->uniform-continuous-distribution, ->uniform-discrete-distribution, %
  ->wishart-distribution, CRP, DP, GP, abs, absorb, acos, asin, atan, %
  bernoulli, beta, binomial, categorical, categorical-crp, categorical-dp, %
  cbrt, ceil, chi-squared, cos, cosh, cov, dirichlet, discrete, exp, %
  exponential, flip, floor, gamma, gen-matrix, log, log-gamma-fn, %
  log-mv-gamma-fn, log-sum-exp, map->CRP-process, map->DP-process, %
  map->GP-process, map->bernoulli-distribution, map->beta-distribution, %
  map->binomial-distribution, map->categorical-crp-distribution, %
  map->categorical-distribution, map->categorical-dp-distribution, %
  map->chi-squared-distribution, map->dirichlet-distribution, %
  map->discrete-distribution, map->exponential-distribution, %
  map->flip-distribution, map->gamma-distribution, map->mvn-distribution, %
  map->normal-distribution, map->poisson-distribution, %
  map->uniform-continuous-distribution, map->uniform-discrete-distribution, %
  map->wishart-distribution, mvn, normal, observe, poisson, pow, produce, %
  rint, round, sample, signum, sin, sinh, sqrt, tag, tan, tanh, transform-sample, %
  uniform-continuous, uniform-discrete, wishart}, %
  sensitive, %
  alsoletter={:,-,+,*,?,/,!}, %
  morecomment=[l];, %
  morestring=[b]", %
  keywordsprefix=[7]:, %
}[keywords,comments,strings]
 

%% file: abstract.tex

In this work we show how to represent policies as programs: that is, as stochastic simulators with tunable parameters.  To learn the parameters of such policies we develop connections between black box variational inference and existing policy search approaches.  We then explain how such learning can be implemented in a probabilistic programming system.  Using our own novel implementation of such a system we demonstrate both conciseness of policy representation and automatic policy parameter learning for a set of canonical reinforcement learning problems.


%% file: introduction.tex

In planning under uncertainty the objective is to find a policy that selects actions, given currently available information, in a way that maximizes expected reward. In many cases an optimal policy can neither be represented compactly nor learned exactly. Online approaches to planning, such as Monte Carlo Tree Search~\cite[]{KS06}, are nonparametric policies that select actions based on simulations of future outcomes and rewards, also known as rollouts. While policies like this are often able to achieve near optimal performance, they are computationally intensive and do not have compact parameterizations. Policy search methods (see \cite{deisenroth_ftr_2011} for a review) learn parameterized policies offline, which then can be used without performing rollouts at test time, trading off improved test-time computation against having to choose a policy parameterization that may be insufficient to represent the optimal policy. 

In this work we show how probabilistic programs can represent parametric policies in a both more general and compact manner. We also develop automatic inference techniques for probabilistic programming systems to do model-agnostic policy search. 
Our proposed approach, which we call black box policy learning (BBPL), is a variant of Bayesian policy search \cite[]{wingate_ijcai_2011,wingate_rep_2013} in which policy learning is cast as stochastic gradient ascent on the marginal likelihood. 

In contrast to languages that target a single domain-specific algorithm \cite[]{andre_aaai_2002,srivastava_uai_2014,nitti_ecml_2015}, our formulation emphasizes the use of general-purpose techniques for Bayesian inference, in which learning is used for inference amortization. To this end, we adapt black-box variational inference (BBVI), a technique for approximation of the Bayesian posterior \cite[]{ranganath_aistats_2014,wingate_arxiv_2013} to perform (marginal) likelihood maximization in arbitrary programs. 
The resulting technique is general enough to allow implementation in a variety of probabilistic programming systems. We show that this same technique can be used to perform policy search under an appropriate planning as inference interpretation, in which a Bayesian model is weighted by the exponent of the reward. The resulting technique, BBPL is closely related to classic policy gradient methods such as REINFORCE \cite[]{williams_ml_1992}. 

We present case studies in the Canadian traveler problem, the RockSample domain, and introduce a setting inspired by Guess Who \cite[]{coster_game_1979} as a benchmark for optimal diagnosis problems.

%% file: background.tex

\section{Policies as Programs}

Probabilistic programming systems \cite[]{milch_chap_2007,goodman_uai_2008,minka_software_2010,pfeffer_rep_2009,mansinghka_arxiv_2014,wood_aistats_2014,GHNR14} represent generative models as programs in a language that provides specialized syntax to instantiate random variables, as well as syntax to impose conditions on these random variables. 
The goal of inference in a probabilistic program is to characterize the distribution on its random variables subject to the imposed conditions, which is done using one or more generic methods provided by an inference backend.




In sequential decision problems we must define a stochastic simulator of an agent, which chooses actions based on current contextual information, and a stochastic simulator of the world, which may have some internal variables that are opaque to the agent, but provides new contextual information after each action. For sufficiently simple problems, both the agent and the world simulator can be adequately described as graphical models. Here we are interested in using probabilistic programs as simulators of both the world and the agent. The trade-off made in this approach is that we can incorporate more detailed assumptions about the structure of the problem into our simulator of the agent, which decreases the size of the search space, at the expense of having to treat these simulators as black boxes from the perspective of the learning algorithm.


In Figure \ref{fig:ctp-overview} we show an example of a program, written in the language Anglican \cite[]{wood_aistats_2014}, which simulates an agent in the Canadian traveler problem (CTP) domain. This agent traverses a graph using depth first search (DFS) as a base strategy, choosing edges either at random, or according to sampled preferences. Probabilistic programs can describe a family of algorithmic policies, which may make use of programming constructs such as recursion, and higher-order functions and arbitrary deterministic operations. This allows us to define structured policies that enforce basic constraints, such as the rule that you should never travel the same edge twice. 

Given a base policy program, we can define different parametrizations that encode additional structure, such as the typical travel distance starting from each edge. We can then formulate a Bayesian approach to policy learning, in which we place a prior on the policy parameters and optimize its hyperparameters to maximize the reward. To do so we employ a planning as inference interpretation \cite[]{toussaint_nc_2006,rawlik_rss_2012,neumann_icml_2011,hoffman2009expectation,hoffman2009new,levine_icml_2013} that casts policy search as stochastic gradient ascent on the marginal likelihood. 

A challenge in devising methods for approximate inference in probabilistic programs is that such methods must deal gracefully with programs that may not instantiate the same set of random variables in each execution. For example, the random policy in Figure~\ref{fig:ctp-overview} will generate a different set of categorical variables in each execution, depending on the path followed through the graph. Similarly, the edge based policy samples values \lsi{(Q u v)} lazily, depending on the visited nodes. 

In this paper we develop an approach to policy learning based on black box variational inference (BBVI) \cite[]{ranganath_aistats_2014,wingate_arxiv_2013}, a technique for variational approximation of the posterior in Bayesian models. We begin by reviewing planning as inference formulations of policy search. We then show how BBVI can be adapted to perform hyperparameter optimization. In a planning as inference interpretation this method, which we call black box policy learning (BBPL), is equivalent to classic policy gradient methods. We then describe how BBPL may be implemented in the context of probabilistic programs with varying numbers of random variables, and provide a language-agnostic definition of the interface between the program and the inference back end.

\section{Policy Search as Bayesian Inference}


In sequential decision problems, an agent draws an action $\u_t$ from a policy distribution $\pi(\u_t \,|\, \x_t)$, which may be deterministic, conditioned on a context $\x_t$. The agent then observes a new context $\x_{t+1}$ drawn from a distribution $p(\x_{t+1} \,|\, \u_t, \x_t)$. In the finite horizon case, where an agent performs a fixed number of actions $T$, resulting in a sequence $\t = (\x_0,\u_0,\x_1,\u_1,\x_2,\ldots,\u_{T-1},\x_T)$, which is known as a trajectory, or roll-out. Each trajectory gets a reward $R(\t)$. Policy search methods maximize the expected reward $J_\q = \E_{p_\q}[R(\t)]$ for a family of stochastic policies $\pi_\q$ with parameters $\q$
\begin{align}
	J_\q
	&=
	\int 
	R(\t)
	p_\q(\t)
	\: d \t
	,
	\\
	p_\q(\t)
	&:= 
	p(\x_0)
	\prod_{t=0}^{T-1}
	\pi(\u_{t} \,|\, \x_{t}, \q)
	\p{\x_{t+1}}{\u_{t},\x_{t}}
	.
\end{align}
We are interested in performing upper-level policy search, a variant of the problem defined in terms of the hyperparameters $\l$ of a distribution $p_\l(\t,\q)$ that places a prior $p_\l(\q)$ on the policy parameters
\begin{align}
	J_\l
	&=
	\int 
	R(\t)
	p_\l(\t,\q)
	\: d \t \, d \q
	,
	\\
	p_{\l}(\t, \q)
	&:= 
	p_\l(\q)
	\p{\t}{\q}
	.
\end{align}
Upper-level policy search can be interpreted as maximization of the normalizing constant $Z_\l$ of an unnormalized density 
\begin{align}
  \label{eq:bbps-density}
  \gamma_{\l}(\t,\q)
  &= 
  p_{\l}(\t,\q) \exp(\beta R(\t))
  ,
  \\
  \label{eq:Zl}
  Z_{\l}
  &=
  \int 
  \gamma_{\l}(\t,\q)
  \:
  d\t
  \,
  d\q
  \\
  &=
  \E_{p_{\l}}[\exp(\beta R(\t))]
  .
\end{align}
The constant $\beta > 0$ has the interpretation of an `inverse temperature' that controls how strongly the density penalizes sub-optimal actions. The normalization constant $Z_{\l}$ is the expected value of the exponentiated reward $\exp(\beta R(\t))$, which is known as the desirability in the context of optimal control \cite[]{kappen_jsmtm_2005,todorov_pnas_2009}. It is not a priori obvious that maximization of the expected reward $J_\l$ yields the same policy hyperparameters as maximization of the desireability $Z_\l$, but it turns out that the two are in fact equivalent, as we will explain in section \ref{sec:methodology}.

In planning as inference formulations, 
$\gamma_\l(\t,\q)/Z_\l$ is often interpreted as a posterior $p_\l(\t,\q \,|\, r)$ conditioned on a pseudo observable $r=1$ that is Bernoulli distributed with probability $p(r=1 \,|\, \t) \propto \exp(\beta R(\t))$, resulting in a joint distribution that is proportional to $\gamma_\l(\t,\q)$,
\begin{align}
	p(r=1,\t,\q)
	&\propto
	p_\l(\t,\q) \exp(\beta R(\t))
	=	
	\gamma_\l(\t,\q).
\end{align}
Maximization of $Z_\l$ is then equivalent to the maximization of the marginal likelihood $p_\l(r=1)$ with respect to the hyperparameters $\l$. In a Bayesian context this is known as empirical Bayes (EB) \cite[]{maritz_mono_1989}, or type II maximum likelihood estimation.

\section{Black-box Variational Inference}

Variational Bayesian methods \cite[]{wainwright_ftml_2008} approximate an intractable posterior with a more tractable family of distributions. For purposes of exposition we consider the case of a posterior $\p{\z,\q}{\y}$, in which $\y$ is a set of observations, $\q$ is a set of model parameters, and $\z$ is a set of latent variables. We write $\p{\z,\q}{\y} = \gamma(\z,\q)/Z$ with
\begin{align}
	\label{eq:bbvi-density}
	\gamma(\z,\q)
	&= 
	\p{\y}{\z,\q}p(\z \,|\, \q)p(\q)
	,
	\\
	Z 
	&=
	\int 
	\gamma(\z,\q)
	\:
	d\z
	\,
	d\q
	.
\end{align}
Variational methods approximate the posterior using a parametric family of distributions $q_\l$ by maximizing a lower bound on $\log Z$ with respect to $\l$
\begin{align}
  \label{eq:bbvi-bound}
  \L_\lambda
  &=
  \E_{q_\l}[\log \gamma(\z,\q) - \log q_\l(z,\q)]
  \\
  &=
  \log Z
  - 
  \Dkl{q_{\l}(\z)}{\gamma(\z)/Z}
  \le
  \log Z
  .
\end{align}
This objective may be optimized with stochastic gradient ascent \cite[]{hoffman_jmlr_2013}
\begin{align}
	\label{eq:bbvi-grad}
	\l_{k+1} 
	&= 
	\l_{k} + \rho_k \nabla_\l \L_\l
	\big|_{\l = \l_k}
	,
	\\
	\nabla_\l \L_\l
	&=
	\E_{q_\l(z)}
	\left[
	\nabla_\l \log q_\l(z)
	\log \frac{\gamma(z,\q)}{q_\l(\z,\q))}
	\right]
	.
\end{align}
Here $\rho_k$ is a sequence of step sizes that satisfies the conditions $\sum_{k=1}^\infty \rho_k = \infty$ and $\sum_{k=1}^\infty \rho_k^2 < \infty$. 
The calculation of the gradient $\nabla_\l \L_\l$ requires an integral over $q_\l$. For certain models, specifically those where the likelihood and prior are in the conjugate exponential family \cite[]{hoffman_jmlr_2013}, this integral can be performed analytically. 

Black box variational inference targets a much broader class of models by  sampling $\z^{[n]},\q^{[n]} \sim q_\l$ and replacing the gradient for each component $i$ with a sample-based estimate \cite[]{ranganath_aistats_2014}
\begin{align}
	\label{eq:bbvi-grad-est}
	\hat{\nabla}_{\l_i} \L_\l
	&=
	\sum_{n=1}^N 
	\nabla_{\l_i} \log q_{\l}(\z^{[n]},\q^{[n]}) (\log w^{[n]} - \hat{b}_i)
	,
	\\
	w^{[n]}
	&=
	\gamma(\z^{[n]},\q^{[n]}) / q_\l(\z^{[n]},\q^{[n]})
	,
\end{align}
in which $\hat{b}_i$ is a control variate that reduces the variance of the estimator
\begin{align}
	\label{eq:bbvi-cv}
	\hat{b}_i
	&= 
	\frac{\sum_{n=1}^N (\nabla_{\l_i} \log q_{\l}(\z^{[n]},\q^{[n]}))^2 w^{[n]}}
	{\sum_{n=1}^N  (\nabla_{\l_i} \log q_{\l}(\z^{[n]},\q^{[n]}))^2}
	.
\end{align}

%% file: methodology.tex




\label{sec:methodology}

The sample-based gradient estimator in BBVI resembles the one used in classic likelihood-ratio policy gradient methods \cite[]{deisenroth_ftr_2011}, such as REINFORCE \cite[]{williams_ml_1992}, G(PO)MDP \cite[]{baxter_rep_1999,baxter_rep_1999b}, and PGT \cite[]{sutton_nips_1999}. 
There is in fact a close connection between BBVI and these methods, as has been noted by e.g.~\cite{dayan1995helmholtz,mnih2014neural} and \cite{ba2014multiple}.

To make this connection precise, let us consider what it would mean to perform variational inference in a planning as inference setting. In this case, we can define a lower bound $\L_{\l,\l_0}$ on $\log Z_{\l_0}$ in terms of a variational distribution $q_\l(\t,\q)$ with parameters $\l$ and an unnormalized density $\gamma_{\l_0}(\t,\q)$ of the form in equation \ref{eq:bbps-density}, with parameters $\l_0$ 
\begin{align}
	\L_{\l,\l_0}
	&=
	\E_{q_\l}[\log \gamma_{\l_0}(\z,\q) - \log q_\l(z,\q)]
  	\\
  	\label{eq:bbpl-bound}
  	&=
	E_{q_\l}
	\left[
	  \beta R(\tau) 
	  + 
	  \log 
	  \frac{p_{\l_0}(\t,\q)}
	  	   {q_{\l}(\t,\q)}
	\right]
\end{align}
If we now choose a variational distribution with the same form as the prior, then $q_\l(\t,\q) = p_{\l_0}(\t,\q)$  whenever $\l = \l_0$. Under this assumption, the lower bound at $\l = \l_0$ simplifies to
\begin{align}
	\label{eq:grad-bbps}
	\L_{\l,\l_0}
	\Big\vert_{\l=\l_0}
	&=
	E_{q_\l}
	\left[
	  \beta R(\tau) 
	\right]
	\Big\vert_{\l=\l_0}
	=
	\beta J_{\l}
	\Big\vert_{\l=\l_0}
	.
\end{align}
In other words, the lower bound $\L_{\l,\l_0}$ is proportional to the expected reward $J_{\l}$ when the variational posterior is equal to the prior. 

The gradient of the lower bound similarly simplifies to
\begin{align*}
	\nabla_\l \L_{\l,\l_0} 
	\Big\vert_{\l=\l_0}
	&= 
	E_{q_{\l}}
	\left[
	  \nabla_\l \log q_\l(\t,\q)
	  \log 
	  \frac{\gamma_{\l_0}(\t,\q)}
	       {q_\l(\t,\q)}
	\right]
	\Big\vert_{\l=\l_0}
	\\
	&=
	E_{q_{\l_0}}
	\left[
	\nabla_\l \log q_\l(\t,\q)
	\Big\vert_{\l=\l_0}
	\beta R(\t)
	\right]
	\\
	&=
	\int d\t d\q ~\nabla_\l q_\l(\t,\q) \Big\vert_{\l=\l_0} \beta R(\t)
	\\
	&=
	\nabla_\l J_\l 
	\Big\vert_{\l=\l_0}
	~.
\end{align*}
The implication of this identity is that we can perform gradient ascent on $J_\l$ by making a slight modification to the update equation
\begin{align}
	\label{eq:bbml-updates}
	\l_{k+1}
	=
	\l_k
	+
	\rho_k
	\hat{\nabla}_\l
	\L_{\l,\l_k} 
	\big|_{\l=\l_k}
	.
\end{align}
The difference in these updates is that instead of calculating the gradient $\hat{\nabla}_\l \L_{\l,\l_0}$ estimate relative to a fixed set of prior parameters $\l_0$, we update the parameters of the prior $p_{\l_k}(\t,\q)$ after each gradient step, and calculate the gradient $\nabla_\l \L_{\l,\l_k}$. We note that the constant $\beta$ is simply a scaling factor on the step sizes $\rho_k$, and will from here on assume that $\beta=1$.

When BBVI is performed using the update step in equation \ref{eq:bbml-updates}, and the variational family $q_\l$ is chosen to have the same form as the prior $p_\l$, we obtain a procedure for EB estimation, which maximizes the normalizing constant $Z_\l$ with respect to the parameters $\l$ of the prior. The difference between the EB and maximum likelihood (ML) methods is that the first calculates the gradient relative to hyperparameters $\l$, whereas the other calculates the gradient relative to the parameters $\q$. Because this difference relates only to the assumed model structure, EB estimation is sometimes referred to as Type II maximum likelihood.

As is evident from equation \ref{eq:grad-bbps}, EB estimation in the context of planning as inference formulations maximizes the expected reward $J_\l$. In the context of a probabilistic programming system this means that we can effectively get three algorithms for the price of one: If we can provide an implementation of BBVI, then this implementation can be adapted to perform EB estimation, which in turn allows us to perform policy search by simply defining models where exponent of the reward takes the place of the likelihood terms. This results in a method that we call black box policy learning (BBPL), which is equivalent to variants of REINFORCE applied to upper-level policy search.

%% file: programs.tex




An implementation of BBVI and BBPL for probabilistic program inference needs to address two domain-specific issues. The first is that probabilistic programs need not always instantiate the same set of random variables, the second is that we need to distinguish between distributions that define model parameters $\theta$ and those that define latent variables $z$, or variables that are part of the context $x$ in the case of decision problems.

Let us refer back to the program in Figure \ref{fig:ctp-overview}. The function \lsi{dfs-agent} performs a recursive loop until a stopping criterion is met: either the target node is reached, or there are no more paths left to try. At each step \lsi{dfs-agent} makes a call to \lsi{policy}, which is created by either calling \lsi{make-random-policy} or \lsi{make-edge-policy}. A random policy samples uniformly from unexplored directions. When depth first search is performed with this policy, we are defining a model in which the number of context variables is random, since the number of steps required to reach the goal state will vary. In the case of the edge policy, we use a memoized function to sample edge preference values as needed, choosing the unexplored edge with the highest preference at each step. In this case the number of parameter variables is random, since we only instantiate preferences for edges that are (a) open, and (b) connect to the current location of the agent. 

As has been noted by \cite{wingate_arxiv_2013}, BBVI can deal with varying sets of random variables quite naturally. Since the gradient is computed from a sample estimate, we can compute gradients for a each random variable by simply averaging over those executions in which the variable exists. Sampling variables as needed can in fact be more statistically efficient, since irrelevant variables that never affect the trajectory of the agent will not contribute to the gradient estimate. BBVI has the additional advantage of having relatively light-weight implementation requirements; it only requires differentiation of the log proposal density, which is a product over primitive distributions of a limited number of types, for which derivatives can be computed analytically. This is in contrast to implementations based on (reverse-mode) automatic differentiation \cite[]{pearlmutter_chap_2008}, as is used in Stan \cite[]{kucukelbir_nips_2015}, which store derivative terms for the entire computation graph.

To provide a language-agnostic definition of BBVI and BBPL, we formalize learning in probabilistic programs as the interaction between a program $\P$ and an inference back end $\B$. The program $\P$ represents all deterministic steps in the computation and has internal state (e.g.~its environment variables). The back end $\B$ performs all inference-related tasks.

A program $\P$ executes as normal, but delegates to the inference back end whenever it needs to instantiate a random variable, or evaluate a conditioning statement. The back end $\B$ then supplies a value for the random variable, or makes note of the probability associated with the conditioning statement, and then delegates back to $\P$ to continue execution. We will assume that the programming language provides some way to differentiate between latent variables $\z$, which are simply to be sampled, and parameters $\q$ for which a distribution is to be learned. In Anglican the syntax \lsi{(sample (tag :policy d))}, as used in Fig.~\ref{fig:ctp-overview}, is used as a general-purpose mechanism to label distributions on random variables. An inference back end can simply ignore these labels, or implement algorithm-specific actions for labeled subsets.

In order for the learning algorithm to be well-defined in programs that instantiate varying numbers of random variables, we require that the each random variable $z_a$ is uniquely identified by an address $a$, which may either be generated automatically by the language runtime, or specified by the programmer. Each model parameter $\theta_b$ is similarly identified by an address $b$.

In BBVI, the interface between a program $\P$ and the back end $\B$ can be formalized with the following rules:
\begin{itemize}
\item Initially $\B$ calls $\P$ with no arguments $\P()$.
\item A call to $\P$ returns one of four responses to $\B$:
\begin{itemize}
	\item[1.] $({\tt sample},a,f, \phi)$: Identifies a latent random variable (not a policy parameter) $\z_a$ with unique address $a$, distributed according to $f_a(\cdot \,|\, \phi_a)$. The back end generates a value $\z_a \sim f_a(\cdot \,|\, \phi_a)$ and calls $\P(\z_a)$.

	\item[2.] $({\tt learn},b,f,\eta)$: For policy parameters, the address $b$ identifies a random variable $\q_b$ in the model, distributed according to a distribution $f_b$ with parameters $\eta_b$. The back end generates $\q_b \sim f_b(\cdot \,|\, \lambda_b)$ conditioned on a learned variational parameter $\lambda_b$ and registers an importance weight $w_b = f_b(\q_{b} \,|\, \eta_b) / f_b(\q_{b} \,|\, \l_b)$. Execution continues by calling $\P(\q_b)$.

	\item[3.] $({\tt factor},c,l)$: Here $c$ is a unique address for a factor with log probability $l_c$ and importance weight $w_c = \exp(l_c)$. Execution continues by calling $\P()$.

	\item[4.] $({\tt return},v)$: Execution completes, returning a value $v$.
\end{itemize}
\end{itemize}
Because each call to $\P$ is deterministic, an execution history is fully characterized by the values for each random variable that are generated by $\B$. However the set of random variables that is instantiated may vary from execution to execution. We write $A, B, C$ for the set of addresses of each type visited in a given execution. The program $\P$ now defines an unnormalized density $\gamma_{\P}$ of the form
\begin{align}
	\gamma_{\P}(\z,\q)
	&:=
	p_{\P}(\z,\q)
	\prod_{c \in C} \exp(l_{c})
	,
	\\
	p_{\P}(\z,\q)
	&:=
	\prod_{a \in A} f_{a}(\z_a \,|\, \phi_{a})
	\prod_{b \in B} f_{b}(\q_b \,|\, \eta_{b})
	~.
\end{align}
Implicit in this notation is the fact that the distribution types $f_a(\cdot \,|\, \phi_a)$ and $f_b(\cdot \,|\, \eta_b)$ are return values from calls to $\P$, which implies that both the parameter values and the distribution type may vary from execution to execution. While $f_a(\cdot \,|\, \phi_a)$ and $f_b(\cdot \,|\, \eta_b)$ are fully determined by preceding values for $\z$ and $\q$, we assume they are opaque to the inference algorithm, in the sense that no analysis is performed to characterize the conditional dependence of each $\phi_a$ or $\eta_b$ on other random variables in the program. 

Given the above definition of a target density $\gamma_{\P}(\z,\q)$, we are now in a position to define the density of a variational approximation $\Q_\l$ to the program. In this density, the runtime values $\eta_b$ are replaced by variational parameters $\lambda_b$
\begin{align}
	\label{eq:Qlambda}
	p_{\Q_\lambda}(z,\q)
	&:=
	\prod_{a \in A} f_{a}(\z_a \,|\, \phi_{a})
	\prod_{b \in B} f_{b}(\q_b \,|\, \lambda_{b})
	~.
\end{align}
This density corresponds to that of a mean-field probabilistic program, where the dependency of each $\theta_b$ on other random variables is ignored.

Repeated execution of $\P$ given the interface described above results in a sequence of weighted samples $(w^{[n]},\q^{[n]},z^{[n]})$, whose importance weight $w^{[n]}$ is defined as
\begin{align}
	\label{eq:bbpl-weights}
	w^{[n]}
	&:=
	\gamma_{\P}(z^{[n]},\q^{[n]}) ~/~
	p_{\Q_\l}(z^{[n]},\q^{[n]})
	\nonumber \\
	&=
	\prod_{b \in B} 
	\frac{f(\theta^{[n]}_b \,|\, \eta_b)}
		 {f(\theta^{[n]}_b \,|\, \lambda_b)}
    \prod_{c \in C}
    \exp l^{[n]}_c
	.
\end{align}



With this notation in place, it is clear that we can define a lower bound $\L_{\Q_\l,\Q_{\l_k}}$ analogous to that of Equation~\ref{eq:bbpl-bound}, and a gradient estimator analogous to that of Equation~\ref{eq:bbvi-grad-est}, in which the latent variables $z$ take the role of the trajectory variables $\tau$. In summary, we can describe a sequential decision problem as a probabilistic program $\P$ in which the log probabilities $l_c$ are interpreted as rewards, parameters $\q_b$ define the policy and all other latent variables $z_a$ are trajectory variables. EB inference can then be used to learn the hyperparameters $\l$ that maximize the expected reward,
as described in Algorithm~\ref{alg:bbpl}.

An assumption that we made when deriving BBPL is that the variational distribution $q_\l(\t,\q)$ must have the same analytical form as the prior $p_{\l_0}(\t,\q)$. Practically this requirement means that a program $\P$ must be written in such a way that the values of the hyperparameters $\eta_b$ have the same constant values in every execution, since their values may not depend on those of random variables. One way to enforce this is to pass $\eta$ as a parameter in the initial call $\P(\eta)$ by $\B$, though we do not formalize such a requirement here.


%

\begin{algorithm}[t]
\caption{Black-box Policy Learning}
\label{alg:bbpl}
\begin{algorithmic}
\State {\bf initialize} parameters $\lambda_{0,b} \leftarrow \eta_b$, iteration $k=0$
\Repeat
\State Set initial $\lambda_{k+1} = \{ \lambda_{k,b} \}_{b \in B}$
\State Run $N$ executions of program $\Q_{\lambda_{k}}$, generating
\State \hspace{1.5em}$(w^{[n]}, \theta^{[n]}, z^{[n]})$ according to Eqns.~\ref{eq:Qlambda}, \ref{eq:bbpl-weights}
\For {each address $b$}
\State Let $N_b \le N$ be the \# of runs containing $b$
\State Let $g_b^{[n]} := \nabla_{\lambda_{k,b}} \log f(\theta_b^{[n]} | \lambda_{k,b})$
\State 
Compute baseline $\hat b_{\lambda_{k,b}}$ from Eq.~\ref{eq:bbvi-cv} 
\State $\hat \nabla_{\lambda_{k,b}} J_{\lambda_k} \leftarrow {N_b}^{-1} \sum g_b^{[n]} (\log w^{[n]} - \hat b_{\lambda_{k,b}})$
\State Update $\lambda_{k+1,b} \leftarrow \lambda_{k,b} + \rho_k \hat \nabla_{\lambda_{k,b}} J_{\lambda_k}$
\EndFor
\State $k \leftarrow k + 1$
\Until parameters $\lambda_b$ converge
\end{algorithmic}
\end{algorithm}

%% file: studies.tex

We demonstrate the use of programs for policy search in three problem domains: (1) the Canadian Traveler Problem, (2) a modified version of the RockSample POMDP, and (3) an optimal diagnosis benchmark inspired by the classic children's game Guess Who. 

These three domains are examples of \emph{deterministic} POMDPs, in which the initial state of the world is not known, and observations may be noisy, but the state transitions are deterministic. Even for discrete variants of such problems, the number of possible information states $x_t = (u_0,o_1,\ldots,u_{t-1},o_t)$ grows exponentially with the horizon $T$, meaning that it is not possible to fully parameterize a distribution $\pi(u \,|\, x, \theta)$ in terms of a conditional probability table $\theta_{x,u}$. In our probabilistic program formulations for these problems, the agent is modeled as an algorithm with a number of random parameters, and we use BBPL to learn the distribution on parameters that maximizes the reward.

We implement our case studies using the probabilistic programming system Anglican \cite[]{wood_aistats_2014}. We use the same experimental setup in each of the three domains. A trial begins with a learning phase, in which BBPL is used to learn the policy hyperparameters, followed by a number of testing episodes in which the agent chooses actions according to a fixed learned policy. At each gradient update step, we use 1000 samples to calculate a gradient estimate. Each testing phase consists of 1000 episodes. All shown results are based on test-phase simulations.

Stochastic gradient methods can be sensitive to the learning rate parameters. Results reported here use a RMSProp style rescaling of the gradient \cite[]{rmsprop_software}, which normalizes the gradient by a discounted rolling decaying average of its magnitude with decay factor $0.9$. We use a step size schedule $\rho_k = \rho_0 / (\tau + k)^\kappa$ as reported in \cite{hoffman_jmlr_2013}, with $\tau=1$, $\kappa=0.5$ in all experiments. We use a relatively conservative base learning rate $\rho_0=0.1$ in all reported experiments. For independent trials performed across a range $1,2,5,10,\ldots,1000$ of total gradient steps, consistent convergence was observed in all runs using over 100 gradient steps. 

The source code for the case studies, as well as the BBPL implementation, is available online.\footnote{\scriptsize{\url{https://bitbucket.org/probprog/black-box-policy-search}}}

\begin{figure}
\begin{center}
	\includegraphics[width=\columnwidth]{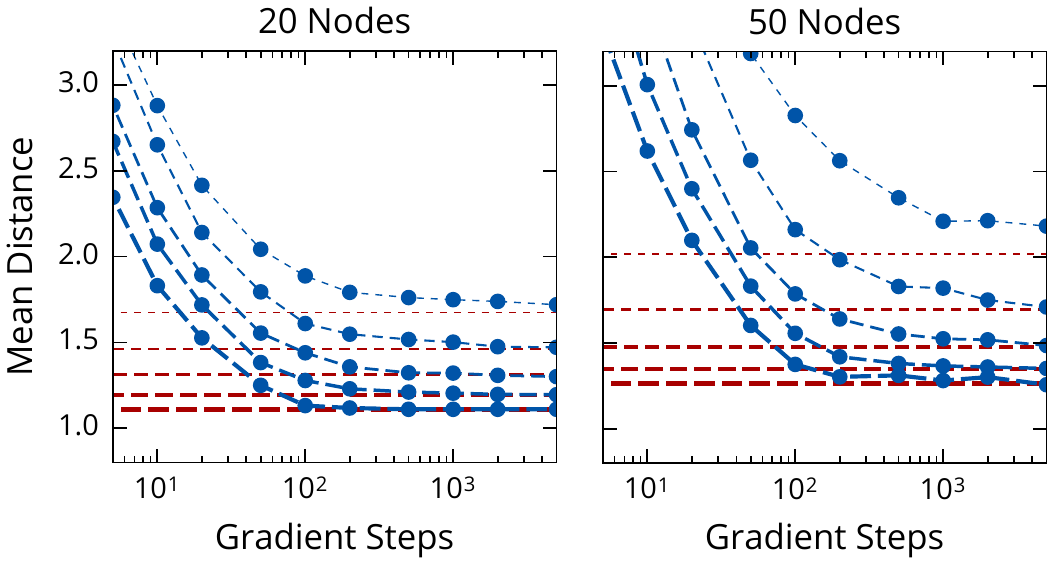}
\end{center}
\caption{\label{fig:ctp-convergence} Convergence for CTP domains of 20 and 50 nodes. Blue lines show the mean traveled distance using the learned policy, averaged over 5 domains. Red lines show the mean traveled distance for the optimistic heuristic policy. Dash length indicates the fraction of open edges, which ranges from 1.0 to 0.6.}
\end{figure}


\subsection{Canadian Traveler Problem}

In the Canadian Traveler Problem (CTP)~\cite[]{papadimitriou_tcs_1991}, an agent must traverse a graph $G=(V,E)$, in which edges may be missing at random. It is assumed the agent knows the distance $d: E \to \R+$ associated with each edge, as well as the probability $p: E \to (0,1]$ that the edge is open, but has no advance knowledge of the edges that are blocked. The problem is NP-hard~\cite[]{fried_tcs_2013}, and heuristic online and offline approaches~\cite[]{eyerich_aaai_2010} are used to solve problem instances.

The results in Figure \ref{fig:ctp-overview} show that the learned policy behaves in a reasonable manner. When edges are open with high probability, the policy takes the shortest path from the start node, marked in green, to the target node, marked in red. As the fraction of closed edges increases, the policy makes more frequent use of alternate routes. Note that each edge has a fixed probability of being open in our set-up, resulting in a preference for routes that traverse fewer edges.  

Figure \ref{fig:ctp-convergence} shows convergence as a function of the number of gradient steps. Results are averaged over 5 domains of 20 and 50 nodes respectively. Convergence plots for each individual domain can be found in the supplementary material. We compare the learned policies against the optimistic policy, a heuristic that selects edges according to the shortest path, assuming that all unobserved edges are open. We observe that mean traveled distance for the learned policy converges to that of the optimistic policy, which is close to optimal.



\begin{figure}
\begin{center}
\includegraphics[width=1.4in]{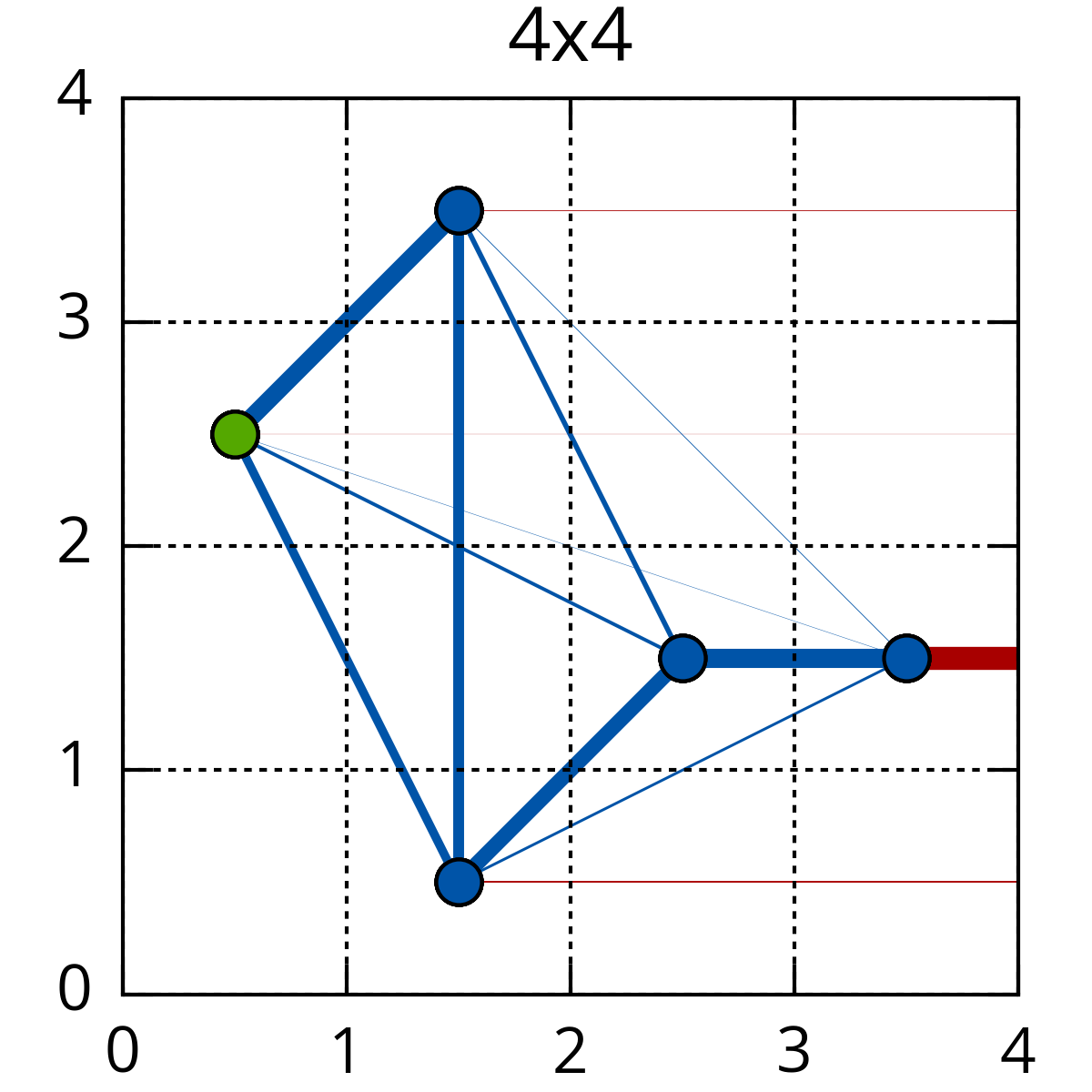}
~~~
\includegraphics[width=1.4in]{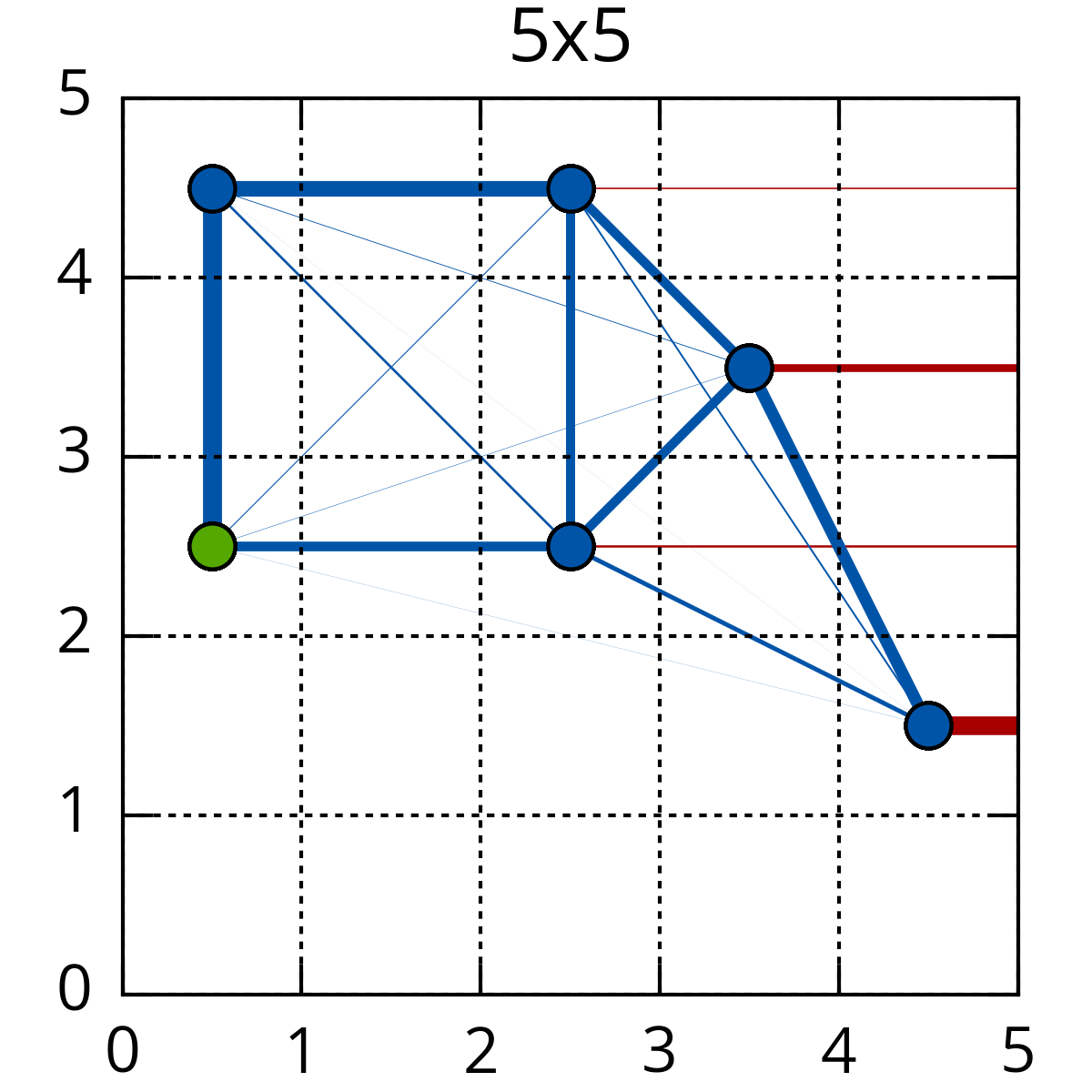}
\\
\vspace{4pt}
\includegraphics[width=1.4in]{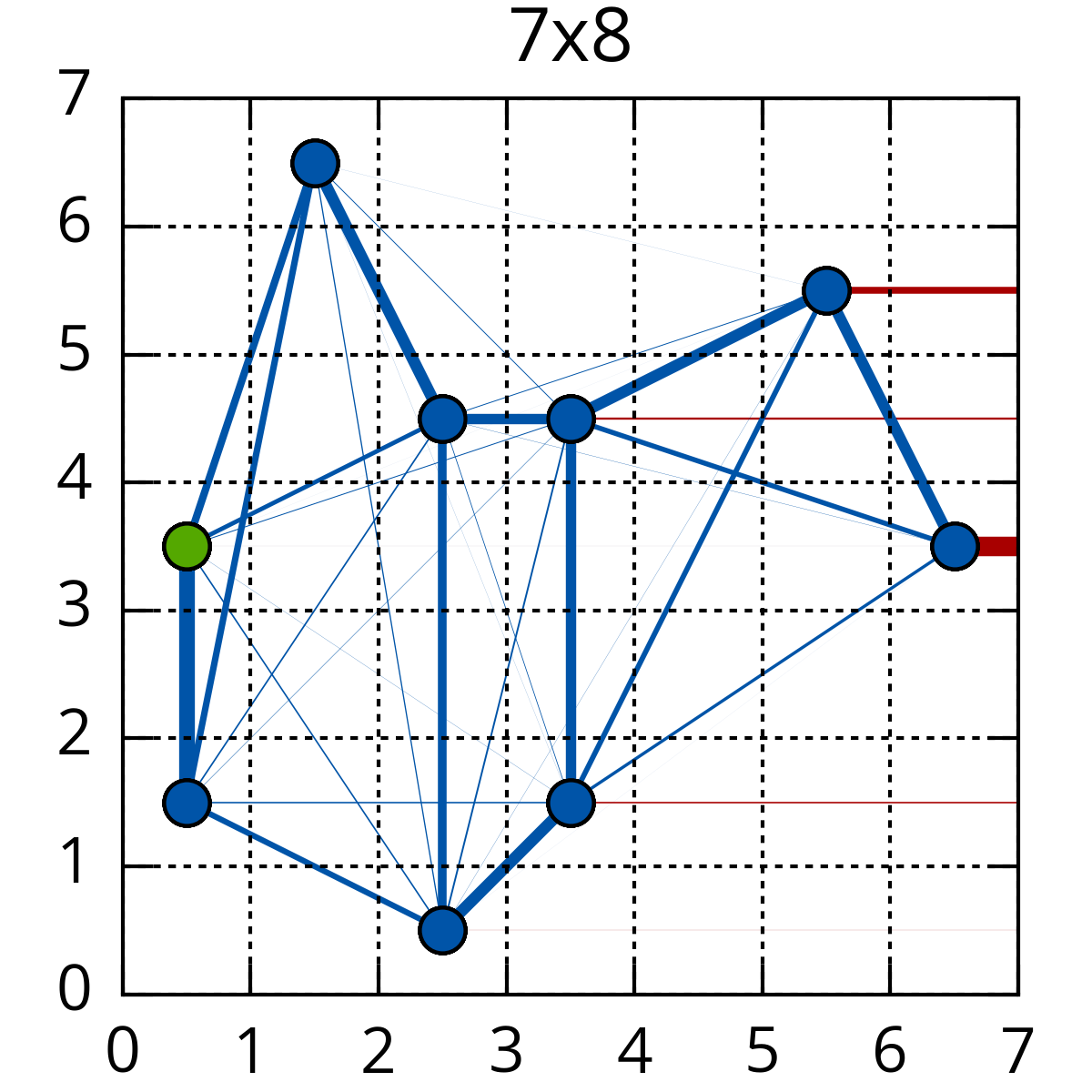}
~~~
\includegraphics[width=1.4in]{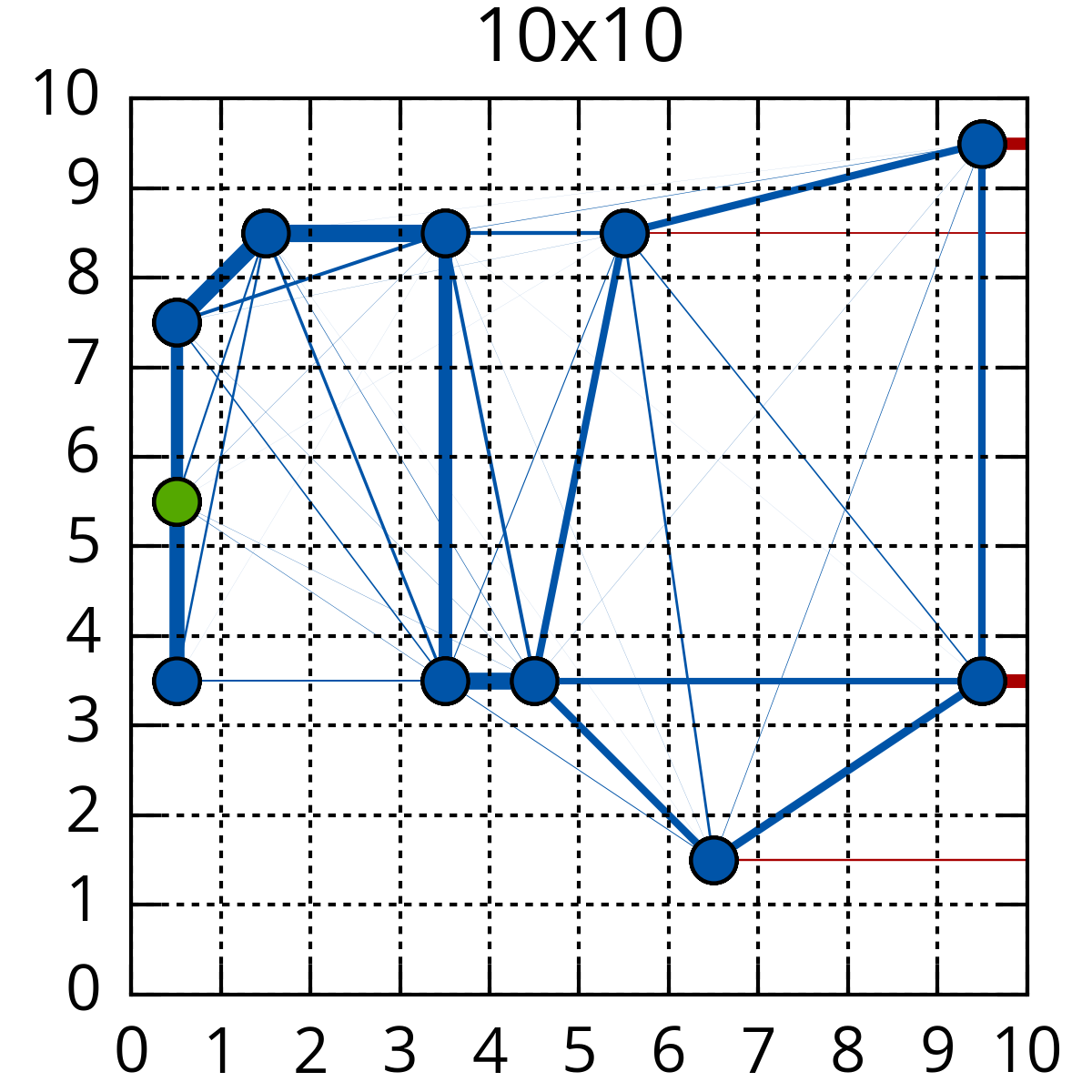}
\end{center}
\caption{\label{fig:rw-graph} Learned policies for the Rock Sample domain. Edge weights indicate the frequency at which the agent moves between each pair of rocks. Starting points are in green, exit paths in red.\vspace{-0.7em}}
\end{figure}

 
\subsection{RockSample POMDP}

In the RockSample POMDP~\cite[]{smith_uai_04}, an $N \times N$ square field with $M$ rocks is given. A rover is initially located in the middle of the left edge of the square. Each of the rocks can be either good or bad; the rover must traverse the field and collect samples of good rocks while minimizing the traveled distance. The rover can sense the quality of a rock remotely with an accuracy decreasing with the distance to the rock. We consider a finite-horizon variant of the RockSample domain, described in the supplementary material, with a structured policy in which a robot travels along rocks in a left-to-right order. 

The policy plots in Figure \ref{fig:rw-graph} show that this simple policy results in sensible movement preferences. In particular we point out that in the $5 \times 5$ instance, the agent always visits the top-left rock when traveling to the top-middle rock, since doing so incurs no additional cost. Similarly, the agent follows an almost deterministic trajectory along the left-most 5 rocks in the $10 \times 10$ instance, but does not always make the detour towards the lower rocks afterwards.

\subsection{Guess Who}

\begin{figure}
\begin{center}
\includegraphics[width=3.1in]{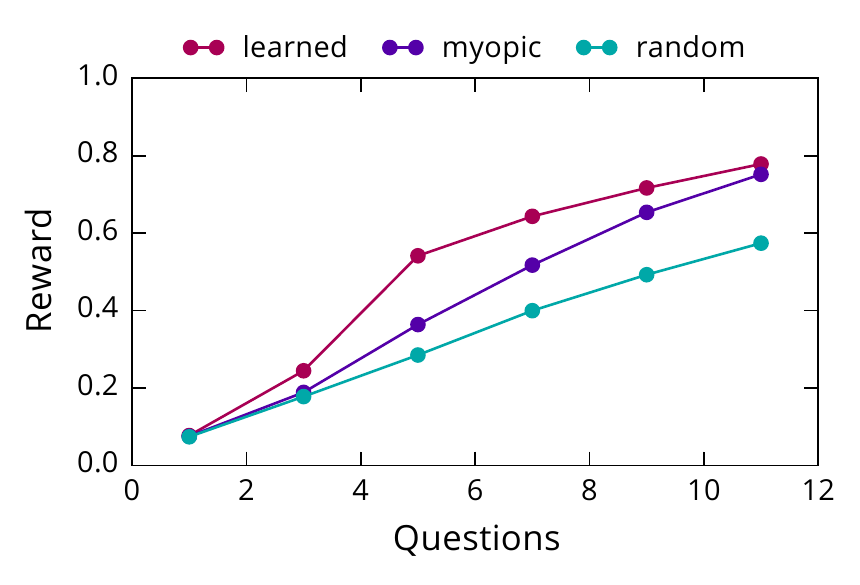}
\vspace{-1.2em}
\end{center}
\caption{\label{fig:guess-who} (left) Average reward in Guess Who as a function of number of questions. (right) Convergence of rewards as function number of gradient steps. Each dot marks an independent restart.\vspace{-0.8em}}
\end{figure}

Guess Who is a classic game in which players pick a card depicting a face, belonging to a set that is known to both players. The players then take turns asking questions until they identify the card of the other player \cite[]{coster_game_1979}.  We here consider a single-player setting where an agent asks a pre-determined number of questions, but the responses are inaccurate with some probability. This is sometimes known as a measurement selection, or optimal diagnosis problem. We make use of  a feature set based on the original game, consisting of 24 individuals, characterized by 11 binary attributes and two multi-class attributes, resulting in a total of 19 possible questions. We assume a response accuracy of 0.9. By design, the structure of the domain is such that there is no clear winning opening question. However the best question at any point is highly contextual. 

We assume that the agent knows the reliability of the response and has an accurate representation of the posterior belief $b_t(s) = \p{s}{ \x_t}$ for each candidate $s$ in given questions and responses. The agent selects randomly among the highest ranked candidates after the final question. We consider 3 policy variants, two of which are parameter-free baselines. In the first baseline, questions are asked uniformly at random. In the second, questions are asked according to a myopic estimate of the value of information \cite[]{hay_uai_2012}, i.e.~the change in expected reward relative to the current best candidates, which is myopically optimal in this setting. Finally, we consider a policy that empirically samples questions $q$ according to a weight $v_q = \gamma^{n_q} (A b)_q $, based on the current belief $b$, a weight matrix $A$, and a discount factor $\gamma^{n_q}$ based on the number of times $n_q$ a question was previously asked. 
Intuitively, this algorithm can be understood as learning a small set of $\alpha$-vectors, one for each question, similar to those learned in point-based value iteration \cite[]{pineau_ijcai_2003}. The discounting effectively ``shrinks'' the belief-space volume associated with the $\alpha$-vector of the current best question, allowing the agent to select the next-best question.

The results in Figure~\ref{fig:guess-who} show that the learned policy clearly outperforms both baselines, which is a surprising result given the complexity of the problem and the relatively simplistic form of this heuristic policy. While these results should not be expected to be in any way optimal, they are encouraging in that they illustrate how probabilistic programming can be used to implement and test policies that rely on transformations of the belief or information state in a straightforward manner.


%% file: discussion.tex

In this paper we put forward the idea that probabilistic programs can be a productive medium for describing both a problem domain and the agent in sequential decision problems. Programs can often incorporate assumptions about the structure of a problem domain to represent the space of policies in a more targeted manner, using a much smaller number of variables than would be needed in a more general formulation. By combining probabilistic programming with black-box variational inference we obtain a generalized variant of well-established policy gradient techniques that allow us to define and learn policies with arbitrary levels of algorithmic sophistication in moderately high-dimensional parameter spaces. Fundamentally, policy programs represent some form of assumptions about what contextual information is most relevant to a decision, whereas the policy parameters represent domain knowledge that generalizes across episodes. This suggests future work to explore how latent variable models may be used to represent past experiences in a manner that can be related to the current information state.

%% file: acknowledgements.tex

\subsection*{Acknowledgements}

We would like to thank Thomas Keller for his assistance with Canadian traveler problem, 
and Rajesh Ranganath for helpful feedback on configuring RMSProp for black-box variational inference.
Frank Wood is supported under DARPA PPAML through the U.S. AFRL under Cooperative Agreement number FA8750-14-2-0006, Sub Award number 61160290-111668.

%% file: appendix.tex


 \section{Anglican}

 All case studies are implemented in Anglican, a probabilistic programming language that is closely integrated into the Clojure language. In Anglican, the macro \lsi{defquery} is used to define a probabilistic model. Programs may make use of user-written Clojure functions (defined with \lsi{defn}) as well as user-written Anglican functions (defined with \lsi{defm}). The difference between the two is that in Anglican functions may make use of the model special forms \lsi{sample}, \lsi{observe}, and \lsi{predict}, which interrupt execution and require action by the inference back end. In Clojure functions, \lsi{sample} is a primitive procedure that generates a random value, \lsi{observe} returns a log probability, and \lsi{predict} is not available. 


 Full documentation for Anglican can be found at 
 \begin{verbatim}
     http://www.robots.ox.ac.uk/~fwood/anglican
 \end{verbatim}

 The complete source code for the case studies can be found at
 \begin{verbatim}
     https://bitbucket.org/probprog/black-box-policy-search
 \end{verbatim}

\section{Canadian Traveler Problem}

\begin{figure}[p]
\begin{center}
	\includegraphics[width=\textwidth]{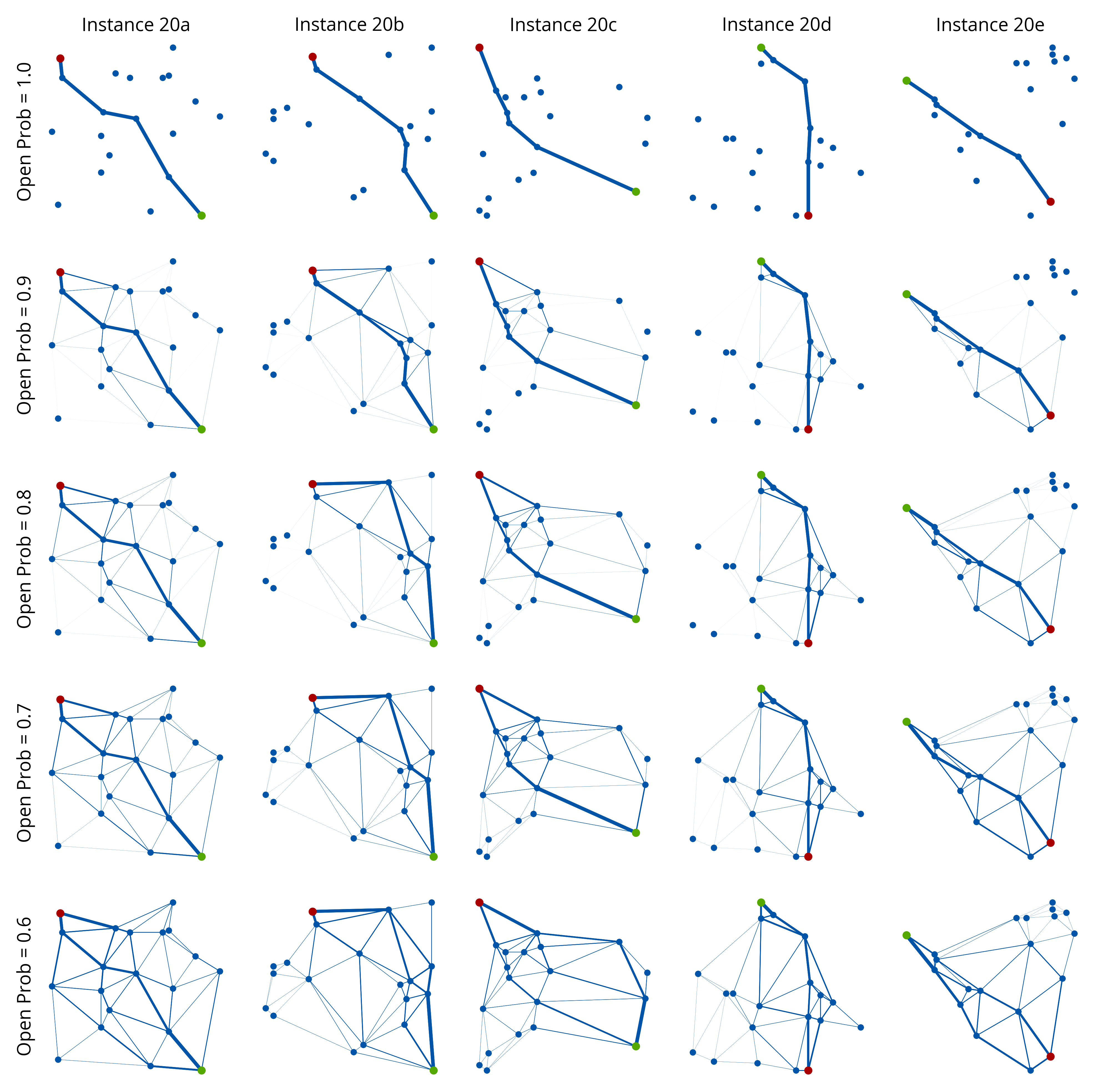}
	\includegraphics[width=\textwidth]{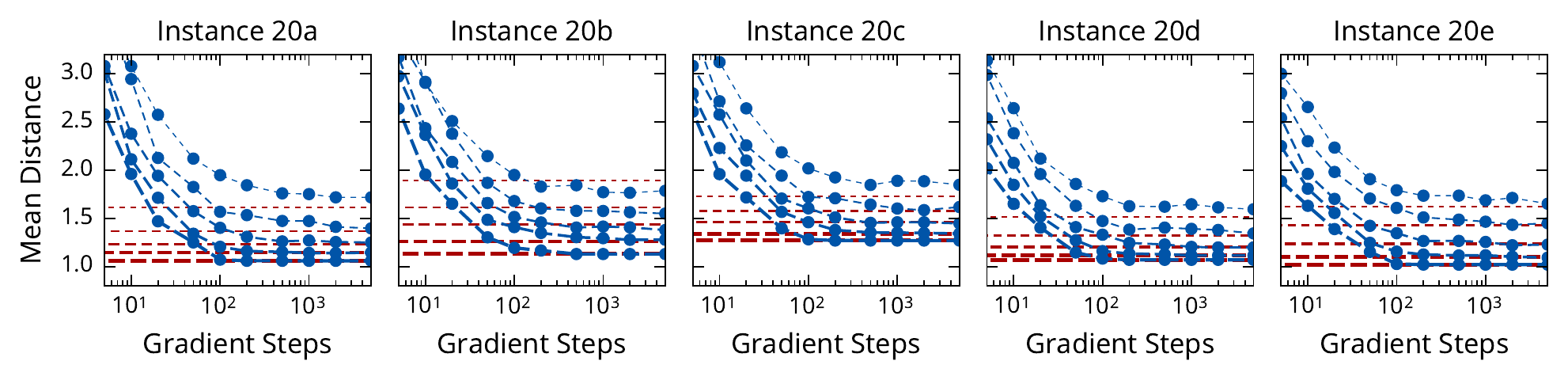}
\end{center}
\caption{\label{fig:ctp-supplementary-20} Canadian traveler problem: edge weights, indicating average travel frequency under the learned policy, and convergence for individual instances with 20 nodes.}
\end{figure}

\begin{figure}[p]
\begin{center}
	\includegraphics[width=\textwidth]{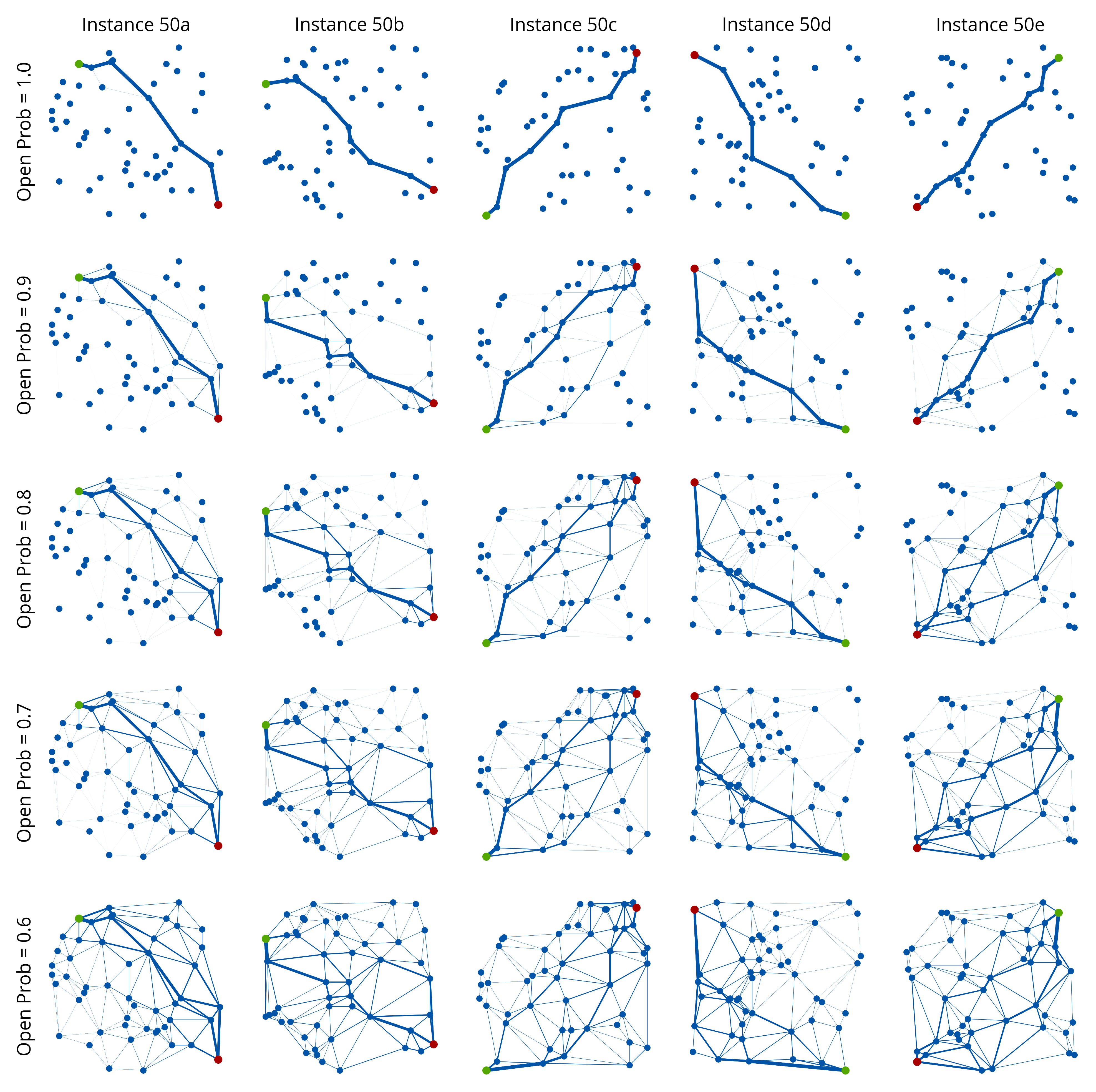}
	\includegraphics[width=\textwidth]{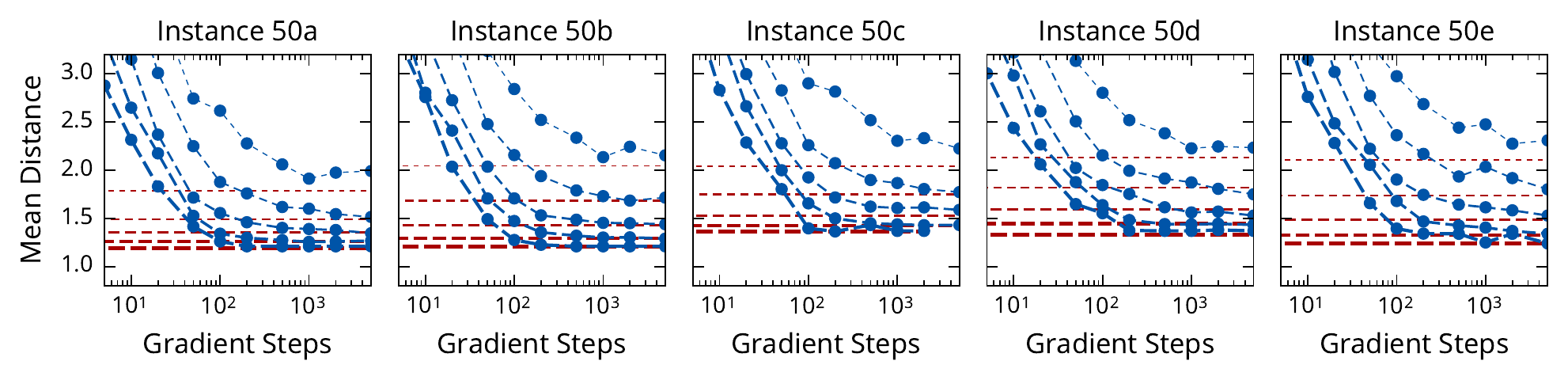}
\end{center}
\caption{\label{fig:ctp-supplementary-50} Canadian traveler problem: edge weights, indicating average travel frequency under the learned policy, and convergence for individual instances with 50 nodes.}
\end{figure}

The complete results for the Canadian traveler problem, showing the performance and convergence for the learned policies
for multiple graphs of different sizes and topologies, 
are presented in Figures~\ref{fig:ctp-supplementary-20}~and~\ref{fig:ctp-supplementary-50}.


\section{RockSample}

The RockSample problem was formulated as a benchmark for value iteration algorithms and is normally evaluated in an infinite horizon setting where the discount factor penalizes sensing and movement. In the original formulation of the problem, movement and sensing incur no cost. The agent gets a reward of 10 for each good rock, as well as for reaching the right edge, but incurs a penalty of -10 when sampling a bad rock. 

Here we consider an adaptation of RockSample to a finite horizon setting. We assume sensing is free, and movement incurs a cost of -1. We structure the policy by moving along rocks in a left-to-right order. At each rock the agent sense the closest next rock and chooses to move to it, or discard it and consider the next closest rock. When the agent gets to a rock, it only samples the rock if the rock is good. The parameters describe the prior over  the probability of moving to a rock conditioned on the current location and the sensor reading. 




 \section{Guess Who}

In Table~\ref{table:ontology} we provide as reference the complete ontology for the Guess Who domain.
At each turn, the player asks whether the unknown individual has a particular value of a single attribute.

 \begin{sidewaystable}
 \begin{tabular}{l|lllllllllllll}
 \hline
 id & beard & ear-rings & eye-color & gender & glasses & hair-color & hair-length & hair-type & hat & moustache & mouth-size & nose-size & red-cheeks\\
 \hline
 alex & false & false & brown & male & false & black & short & straight & false & true & large & small & false\\
 alfred & false & false & blue & male & false & ginger & long & straight & false & true & small & small & false\\
 anita & false & false & blue & female & false & blonde & long & straight & false & false & small & small & true\\
 anne & false & true & brown & female & false & black & short & curly & false & false & small & large & false\\
 bernard & false & false & brown & male & false & brown & short & straight & true & false & small & large & false\\
 bill & true & false & brown & male & false & ginger & bald & straight & false & false & small & small & true\\
 charles & false & false & brown & male & false & blonde & short & straight & false & true & large & small & false\\
 claire & false & false & brown & female & true & ginger & short & straight & true & false & small & small & false\\
 david & true & false & brown & male & false & blonde & short & straight & false & false & large & small & false\\
 eric & false & false & brown & male & false & blonde & short & straight & true & false & large & small & false\\
 frans & false & false & brown & male & false & ginger & short & curly & false & false & small & small & false\\
 george & false & false & brown & male & false & white & short & straight & true & false & large & small & false\\
 herman & false & false & brown & male & false & ginger & bald & curly & false & false & small & large & false\\
 joe & false & false & brown & male & true & blonde & short & curly & false & false & small & small & false\\
 maria & false & true & brown & female & false & brown & long & straight & true & false & small & small & false\\
 max & false & false & brown & male & false & black & short & curly & false & true & large & large & false\\
 paul & false & false & brown & male & true & white & short & straight & false & false & small & small & false\\
 peter & false & false & blue & male & false & white & short & straight & false & false & large & large & false\\
 philip & true & false & brown & male & false & black & short & curly & false & false & large & small & true\\
 richard & true & false & brown & male & false & brown & bald & straight & false & true & small & small & false\\
 robert & false & false & blue & male & false & brown & short & straight & false & false & small & large & true\\
 sam & false & false & brown & male & true & white & bald & straight & false & false & small & small & false\\
 susan & false & false & brown & female & false & white & long & straight & false & false & large & small & true\\
 tom & false & false & blue & male & true & black & bald & straight & false & false & small & small & false\\
 \end{tabular}
 \caption{Ontology for the Guess Who domain, consisting of 24 individuals, characterized by 11 binary attributes and two multi-class attributes.}
 \label{table:ontology}
 \end{sidewaystable}







